\documentclass[acmsmall, sigconf,screen]{acmart}

\renewcommand\footnotetextcopyrightpermission[1]{} 
\usepackage[T1]{fontenc}
\usepackage{nopageno}
\usepackage{graphicx} 
\usepackage{pgfplots}
\usepackage{amsmath,amssymb}
\DeclareMathOperator{\E}{\mathbb{E}}
\usepackage{breqn}
\usepackage{dsfont}
\usepackage{amsthm}
\usepackage{hyperref}
\usepackage[utf8]{inputenc}
\usepackage[english]{babel}
\usepackage{mathtools}

\usepackage[boxed]{algorithm2e}

\newcounter{l2}
\newcommand{\balphlist}{\begin{list}{(\alph{l2})}{\usecounter{l2}}}
\newcommand{\barablist}{\begin{list}{\arabic{l1}}{\usecounter{theorem}}}
\newcounter{foo}
\newcounter{theorem}

\usepackage{setspace} 
\usepackage{array} 
\usepackage{paralist} 
\usepackage{verbatim} 
\usepackage{subfig} 
\usepackage{amsthm} 
\usepackage{amssymb}
\usepackage{mathtools}
\usepackage{todonotes}

\usepackage{bbm}
\usepackage{fancyhdr} 
\usepackage[nottoc,notlof,notlot]{tocbibind} 
\usepackage[titles,subfigure]{tocloft} 

\usepackage{ marvosym }

\theoremstyle{definition}

\DeclareUnicodeCharacter{2212}{-}
\DeclareMathOperator*{\argmax}{arg\,max}

\DeclareMathOperator*{\expect}{\E}

\pgfplotsset{compat=1.17}
\usepackage{algorithmic}

\begin{document}

\title[Offline-Online Reinforcement Learning for Office Demand Response]{Offline-Online Reinforcement Learning for Energy Pricing in Office Demand Response: Lowering Energy and Data Costs}

\settopmatter{authorsperrow=1} 
\newcommand{\tsc}[1]{\textsuperscript{#1}} 
\author{Doseok Jang\tsc{1}, Lucas Spangher \tsc{1}, Manan Khattar \tsc{1}, Utkarsha Agwan \tsc{1}, Selvaprabuh Nadarajah \tsc{2}, Costas Spanos \tsc{1}} 
\affiliation{
  \institution{ 1. Department of Electrical Engineering and Computer Sciences, University of California, Berkeley}
  \city{Berkeley}
  \state{California}
  \country{USA}
  \institution{2. Department of Information and Decision Sciences, University of Illinois, Chicago}
}

\begin{abstract}
Our team is proposing to run a full-scale energy demand response experiment in an office building. Although this is an exciting endeavor which will provide value to the community, collecting training data for the reinforcement learning agent is costly and will be limited. In this work, we examine how offline training can be leveraged to minimize data costs (accelerate convergence) and program implementation costs. We present two approaches to doing so: pretraining our model to warm start the experiment with simulated tasks, and using a planning model trained to simulate the real world's rewards to the agent. We present results that demonstrate the utility of offline reinforcement learning to efficient price-setting in the energy demand response problem.  
\end{abstract}



\maketitle

 \section{Introduction} \label{sec:intro}
 
The bridge from simulation to experiment is difficult to cross for both statistical and practical reasons. When considering experiments of phenomena in the energy grid, techniques to help a learned controller retain some information gained during simulation can be very beneficial for reducing these practical considerations.  

We focus here on the electrical grid. As the grid decarbonizes, volatile resources like wind and solar will replace on-demand resources like fossil fuels, and there arises a mismatch between generation and demand. Grids that do not prepare for this question will face daunting consequences, from curtailment of resources \citep{spangher2020prospective} to voltage instability and physical damage, despite having adequate generative capability. Indeed, these problems will only grow larger as the world moves away from fossil fuels.  
 
 Demand response, a strategy in which customers are incentivized to shift their demand for energy resources to parts of the day where generation is plentiful by manipulating energy prices, is seen as a common solution to the problem of generation volatility. Given the lack of needed material infrastructure and cheapness of the incentives, it has several positives above physical energy storage systems. 
 
 Building energy is a primary target of demand response, and both the central administration of signals and building-level response has been thoroughly studied in residential and industrial settings (\citep{asadinejad2018evaluation}, \citep{ma2015cooperative}, \citep{li2018integrating}, \citep{yoon2014dynamic}, \citep{johnson2015dynamic}.) However, while physical infrastructures\citep{das2020occupants} of office buildings have been studied for demand response (\citep{8248801}), there has been no large scale experiment aimed at eliciting a behavioral demand response using prices. The lack of an office-centered study is understandable when we consider that most offices do not have a mechanism to pass energy prices onto workers. If they did, however, not only would a fleet of decentralized batteries -- laptops, cell phone chargers, etc. be able to be coordinated to function as a large deferable resource, but building managers could save money by adapting their buildings' energy usage to a dynamic utility price.

The SinBerBEST collaboration has developed a Social Game\citep{konstantakopoulos2019design} that facilitates workers to engage in a competition around energy \citep{konstantakopoulos2019deep}, \citep{das2019novel}. Through this framework, a first-of-its-kind experiment has been proposed to implement behavioral demand response within an office building \citep{spangher2020prospective}. Prior work has proposed to describe an hourly price-setting controller that learns how to optimize its prices \citep{spangher2020augmenting} to maximize efficient energy usage by workers. However, the use of an AI price-setting controller gives rise to a tradeoff between \textit{energy cost} and \textit{data cost}. \textit{Energy cost} is the price of the energy used by workers. \textit{Data cost} is the number of days the price controller must be deployed to learn a policy that is profitable compared a reasonable baseline. A fully trained price controller has high data cost (on the order of decades), high energy cost in the short run, and low energy cost in the long run. Currently, this high initial data and energy cost are the most significant hurdle to the deployment of AI price-setting in energy demand response. Given the costliness of data in this experiment and the work that has been put into building a complex simulation environment, an offline-online approach -- warm-starting the experiment's controller with learning from offline simulations -- could prove valuable to its success.  

We report experiments with offline-online reinforcement learning, as well as a DAgger inspired approach to mixing offline and online data. We will in Section \ref{sec:models} explain the models that underly our simulation of the Social Game. In Section \ref{sec:methods} we will contextualize the architecture of our price controller within reinforcement learning, explain the motivation of our planning environment, and describe how we test our warm-started controller. In Section \ref{sec:results} we will give results. Finally, in Section \ref{sec:Discussion} we will discuss implications of the controller and the future work this entails.

 \section{Models} \label{sec:models}

\subsection{Price Setting Problem}

We consider an office of 500 simulated people with 10 hour workdays. Each simulated person $i$ has a baseline energy expenditure $\vec{b}_i = [b_1,...,b_{10}]_i$ and are aware of hourly prices set by a controller $\vec{p} = [p_1,...,p_{10}]$. Each person consumes $\vec{d}$ energy deterministically with respect to prices, i.e.  $\vec{d}_i = f_i(\vec{b}_i, p)$, described below in Section \ref{sec:Environment}. The building manager implements a price-setting controller's in order to minimize people's total energy cost, defined by $\sum_i{\vec{g}^T\vec{d}_i}$, where $\vec{g}$ are TOU grid prices.  There are numerous challenges in solving the model, with the main ones being setting a price that accounts for heterogeneity of response function $f$, figuring out the difference between non-deferable load and deferable, and non-linearities in price sensitivity. All show up in reinforcement learning as difficulties in learning rate. 

\subsection{Social Game}

The Social Game is administered as described in \citep{konstantakopoulos2019design}; office workers compete with each other to have the lowest cost of energy according to the controller's prices. Each receives a default number of points for each hour they are in the office, scaled to their historical average. Players compete on a per-round basis, where each round lasts two weeks. The points they spend on energy -- i.e. $\vec{d}_i^T\vec{g}$ are subtracted from the default totals. Thus, players have the chance to accumulate points throughout each round by playing in a way that reduces their overall cost of energy below a baseline computed from their historical usage. Every two weeks, the top 33\% of players are entered into a prize pool where winners are selected randomly according to Vicky-Clark-Groves (VCG) mechanism of auction design. Approximately \$400 in prizes are given out. Thus, the total cost to a building owner is \$800 a month, which we measure our pricing schemes against as beating. The Social Game investment is reasonable considering that relative to the proposed \$400, a medium-sized office building of 500 people may run up energy costs in computation, ventilation, and air conditioning on the order of \$10k every two weeks, with desk-level energy expenditures accounting for roughly \$1-2k \citep{konstantakopoulos2017robust}. 

\section{Social Game Simulation Environment}\label{sec:Environment}
 
We summarize an OpenAI gym environment built to model the Social Game and instantiate the price-setting problem in office buildings \cite{spangherofficelearn}.  Each step in the environment is a day, where the agent proposes prices to office workers. Notably, we employ several different models of simulated response, with two levels of complexity: ``Deterministic Function'' and ``Curtail and Shift''. Their descriptions are listed below:

\subsection{``Deterministic Function'' Person}

We include three types of deterministic response within one type of agent, with the option of specifying a mixed office composed of all three types. 

A "Deterministic Function" Person with \textbf{linear response} decreases their energy consumption linearly below an average historical energy consumption baseline. Therefore, if $m$ is a simulation set points multiplier, the energy demand is $ \vec{d} = \vec{b} - \vec{p} * m$, clipped at ceiling and floor values $d_{min}$ and $d_{max}$, which are 5\% and 95\% of a historical energy distribution from data in \citep{spangher2019visualization}. A "Deterministic Function" Person with \textbf{sinusoidal response} is one who responds to points towards the middle of the distribution and not well to low or high points. Therefore, the energy demand $\vec{d}$ is $ \vec{d} = \vec{b} - \sin{\vec{p}} * m$, clipped at $d_{min}$ and $d_{max}$.

A "Deterministic Function" Person with \textbf{threshold exponential response}, we define an office worker who does not respond until points pass a threshold, at which point they respond exponentially. Therefore, the energy demand $d$ is  $\vec{d} = \vec{b} - (\exp{\vec{p}} * \mathds{1}( \vec{p} > 5))$ , clipped at $d_{min}$ and $d_{max}$. 
 
\subsection{``Curtail And Shift Office Worker''}

Office workers need to use electricity to do their work, and may be unable to curtail their load below a minimum threshold, e.g. the power needed to run a PC. They may have the ability to shift their load over a definite time interval, e.g. choosing to charge their laptops ahead of time or at a later time. We model a response function that exhibits these behaviors. We can model the aggregate load of a person ($\vec{d}$) as a combination of fixed demand ($d^{fixed}$), curtailable demand ($d^{curtail}$), and shiftable demand ($d^{shift}$), i.e., $\vec{d} = d^{fixed} + d^{curtail} + d^{shift}$. All of the curtailable demand is curtailed for the $T_{curtail}$ hours (set to $3$ hours in practice) with the highest points, and for every hour $t$ the shiftable demand is shifted to the hour within $[t - T_{shift}, t+T_{shift}]$ with the lowest energy price. For example, such an office worker may need to charge their appliances for a total of 1000 Wh throughout the day, 300 of which are for printing documents that could be curtailed, 300 of which are for presenting at a meeting whose time can be shifted, and 400 of which are the minimum required to run a PC. Upon receiving a price signal with high prices from 11am-2pm, this simulated worker would curtail their printing away from the 3 hours with the highest energy price, and schedule their meeting at the hour within $[t - T_{shift}, t + T_{shift}]$ with the lowest energy price. This would decrease their energy usage.

\begin{figure}
\centerline{\includegraphics[width=\linewidth]{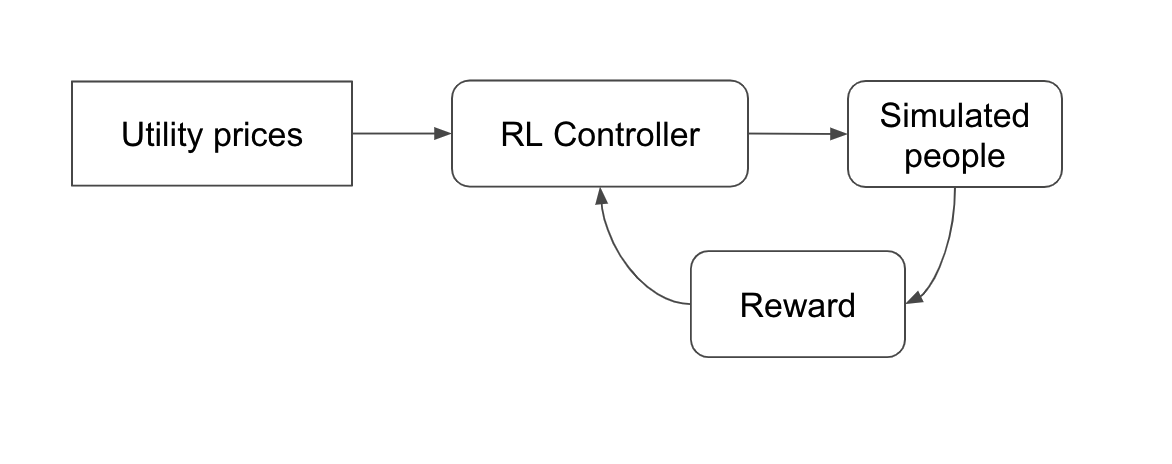}}
\caption{Reinforcement Learning Control Flow} \label{fig:RLcontroller}
\end{figure}
  
 \section{Reinforcement Learning for Smart Energy Pricing} \label{sec:methods}
 Reinforcement learning (RL) is a type of agent-based machine learning where control of a complex system requires actions that optimize the system \citep{sutton2018reinforcement}, i.e. they seek to optimize the expected sum of rewards for actions ($a_t$) and states ($s_t$) in a policy parameterized by $\theta$; i.e.,  $J(\theta) = \E(\sum_{s_t,a_t \sim p_{\pi}}[r(s_t, a_t)])$. It naturally fits the price setting problem, as the RL controller's reward is a transformation of the building manager's objective.   Often we model a policy as a deep neural network with weights $\theta$ that takes states as input and outputs actions. RL is useful in contexts where actions and environments are simple or data is plentiful, with early use cases being optimizing the control of backgammon \citep{tesauro1994td}, the cart-pole problem, and Atari \citep{mnih2013playing}. 
 
  RL has been applied to a number of demand response situations, but almost all work centers on agents that directly schedule resources \citep{6963416}, \citep{7018632}, \citep{6848212}, \citep{6915886}, \citep{RAJU2015231}, \citep{FUSELLI2013148}. RL architectures can vary widely, for example Kofinas et. al. deploy a fuzzy Q-learning multi-agent that learns to coordinate appliances to increase reliability \citep{KOFINAS201853}. In another example, Mbuwir et. al. manage a battery directly using batch Q-learning \citep{en10111846}. 
  
  We seek to use an RL agent to solve the price setting problem; can an RL agent preemptively estimate the most effective demand response price using historical data and implicitly predicting causal factors? Furthermore, can we pretrain an agent in simulation that can quickly adapt to real-world data? We employ Soft Actor Critic (SAC), pretraining, and a planning model to train an agent in OfficeLearn to optimize our policy network to adapt quickly to new environments.
  
  \subsection{Online Reinforcement Learning}
  \subsubsection{SAC}
  RL architectures may be grouped in families between which  learning differs significantly; one such family is the family of Actor-Critic architectures. These methods train two neural networks: the Actor and the Critic. The Actor takes as input the state of the environment and outputs probabilities for each action in its action space; it maps states to actions. The Critic takes as input the state of the environment and returns an estimate of the rewards the model is expected to attain in the future; it maps states to state values. The Critic is used to guide the training of the Actor, resulting in more performant training. We use Soft Actor Critic (SAC) \citep{haarnoja2018soft} in this paper, a variant that adds an entropy term to the reward as a way of encouraging exploration, changing the RL problem to:
\begin{equation}
    \pi^* = \argmax_\pi \expect_{\tau \sim \pi}[\sum_{t=0}^{\infty}\gamma^t(R(s_t, a_t, s_{t+1})+\alpha H(\pi(\cdot | s_t)))]
 \end{equation} where $\alpha$ is the weight given to the entropy regularization and $H$ computes the entropy of the action probability distribution.
  SAC is considered one of the state of the art RL algorithms for continuous action spaces like ours.

  We use the RLLib \cite{liang2018rllib} implementation of SAC, and their default neural network architecture of two hidden layers with 256 units each and $\tanh$ activations. The reward for the price-setting agent is $-\log(d^tg) - \lambda * \mathds{1}
(d^tg - \hat{d})$, where $d$ is the demand of the person it studies, $g$ is the grid pricing, $\hat{d}$ is minimal baseline energy demand that the grid must meet, and $\lambda * \mathds{1}
(d^tg - \hat{d})$ is a regularizing penalty that is applied if the simulated energy grid does not meet $\hat{d}$. The penalty helps avoid the locally optimal solution of the agent driving down prices indiscriminately without regard for energy supply. For other implementation choices, please see our Github\footnote{Please find our Github here: \url{https://github.com/Aphoh/temp_tc/tree/planning_dagger2}} and RLLib.

The naive approach to using SAC to learn a price-setting controller would be to use purely online data: deploying the model in the real world, recording states and energy costs (which are used to calculate rewards), and using only this real world data to train the controller. We will refer to this approach as "Online SAC"
  
  \subsection{Offline-Online Reinforcement Learning}
  With the online "vanilla" SAC optimization procedure, several decades worth of real-world training data would have to be collected to fully train an hourly price-setting controller \citep{spangher2020augmenting} in our Social Game.  We seek to leverage a detailed simulation with behaviorally reasonable dynamics encoded in a model that can train on both simulated and experimental environments to accelerate this process. SAC is an off-policy algorithm, which means that it can be trained on state transitions that did not originate from its policy. This allows SAC to be used to train networks on offline datasets of previously collected samples. Thus we propose pretraining SAC on an offline dataset of state transitions collected from our Social Game simulations, in order to learn a warm-started neural network initialization that can generalize to the experimental environment with few real world steps, decreasing data cost. We will refer to this procedure as "Offline-Online SAC", as this SAC is first trained on offline data before transitioning to online data.

\subsection{Dataset Aggregation (DAgger)}
Instead of the strict transition from offline to online training in Offline-Online SAC, we explore interleaving offline and online training through a DAgger inspired weighting scheme. For our "offline" component in this variant, we explore the possibility of using a planning model to accelerate training. This model is a neural network trained to predict the responses of people to a proposed price, essentially a trained simulation of the rewards an agent would receive given a state and an action in the real world. We use this planning model as our offline component here instead of the Social Game simulations from Section \ref{sec:Environment}, because we believe training on data from two completely different distributions at the same time would not yield a model that learns efficiently for the real world task. We thus try to make our offline data source as close as possible to the online data. However, with a limited number of samples it is impossible to train a planning model that exactly predicts the reward from the real environment, so we are still faced with the issue of having a source of training data that may not be aligned to the distribution of test data.

DAgger \citep{ross2011reduction} is a meta-algorithm that helps solve the problem of distribution shift between training and test data in imitation learning. In the original paper, DAgger was used to solve the problem of training on one distribution (states reached by humans) and testing on another (states reached by the RL agent). Inspired by this, we attempt to adapt DAgger to bridge the gap between two distributions of training data: samples from the target environment, and samples from our planning model. 

In order to mix data from the planning model and target environment, we employ a weighting strategy inspired by DAgger. We alternate training in the planning model and target environment, exponentially decaying the ratio of planning model steps as training continues. Our rationale is that our RL price controller should glean as much information as possible from the planning model first, since sampling from the planning model has negligible cost compared to sampling steps from the target environment. Once the model has learned enough from the planning model, steps from the target environment are slowly introduced into the training dataset, ultimately producing a price controller that performs well on the target environment with fewer steps. In this way, we dynamically weight the two data sources for SAC for more efficient learning. For our experiments, we set the initial ratio of planning steps to target environment steps as $M=10$, and exponential decay parameter $\beta=0.99$. The algorithm for our data mixing procedure can be seen in Alg. \ref{alg:DAgger}. We refer to this training procedure as "DAgger SAC" since it interleaves online real world and offline planning model data to form an aggregated dataset for SAC to optimize the price controller. We will also refer to "Offline-DAgger SAC", which consists of employing DAgger SAC during Offline-0nline SAC's online portion. Though DAgger SAC does have an upfront data cost to train the planning model, our results show that this algorithm does ultimately decrease data cost by leveraging knowledge from the planning model.

\begin{algorithm}[h]
\caption{Planning model and target environment data mixing procedure}
\label{alg:DAgger}
\begin{algorithmic}
   \STATE Initialize $D \longleftarrow 0$
   \STATE Initialize $\pi_1$ to any policy in $\prod$
   \FOR{$i=1$ {\bfseries to} $N$}
   \STATE Sample $T$-step trajectories using $\pi_i$
   \FOR{$j=1$ {\bfseries to} $\lfloor M_i \rfloor$ }
       \STATE Get dataset $D_{ij}={(s, \pi_i(s), R^*(s))}$ of visited states and actions taken by $\pi_i$, and rewards given by the planning model.
   \ENDFOR
   \STATE Get dataset $D_{i0} = {(s, \pi_i(s), R(s))}$ of visited states and actions taken by $\pi_i$, and rewards given by the target environment.
   \STATE Aggregate datasets: $D \longleftarrow D \bigcup {D_{i0}, D_{i1}...D_{iM}}$
   \STATE Train policy $\pi_{i+1}$ on $D$
   \STATE Let $M_{i+1} = M_i * \beta$
   \ENDFOR 
   
\end{algorithmic}
\end{algorithm}

\subsection{Time of Use (TOU) and Flat Controls}
As baselines to assess the utility of our RL agent, we introduce two control price setting algorithms: TOU and Flat Pricing.

For our TOU control, the same TOU price signal that is used in the RL agent's reward, i.e. \$[.09, .09, .09, .39, .39, .39, .09, .09, .09, .09] over a 10 hour work-day, is simply passed onto the simulated office workers as a static price signal. Each office worker would see that at hour 0, the price would be 0.09 \$ per kWh of energy used, at hour 3 the price would be 0.39 \$, etc. and change their behavior accordingly. The prices for TOU stem from the assumption that energy use will be much cheaper during certain parts of the day than others, e.g. wind and solar power will produce more energy during hours when the wind is blowing and the sun is out. This should shift energy spending behaviour toward these hours with cheaper and more plentiful energy, ultimately resulting in cheaper energy costs compared to a flat price signal. 

Similarly to TOU, a flat price signal is passed to the simulated office workers to estimate simulation behavior under no price signal. As the output is invariant to the value of the signal, we use [0, 0, ..., 0] as the price signal. Under this price signal, the simulated office workers behave as they would with no demand response in place at all; since all prices are uniform, they do not adjust their behavior in any way from their normal energy use. 

Note that in the real world,  a flat pricing strategy would result in some effect, as office workers would compete for general energy reduction. Indeed, the SinBerBEST collaboration has run experiments for generalized energy reduction \citep{spangher2019visualization}. However, as this is not the subject of the paper and because the general pricing in offices is no price signal at all, we formulate our simulation such that the office workers respond a flat signal as they would to no signal at all, to provide an accurate control. Additionally, we note that a flat pricing signal is inconvenient from the standpoint of a building manager's financial incentives, as it is less likely to generate a profit than TOU pricing given underlying energy demands may not line up with utility pricing.   

\begin{table}[h]
\label{comp-table}
\vskip 0.15in
\begin{center}
\begin{small}
\begin{sc}
\begin{tabular}{lcccr}
\toprule
Social Game  & No Social Game \\
\midrule
RL Price Controllers & No pricing \\ 
TOU Pricing &  Flat Pricing$^*$ \\
\bottomrule
\end{tabular}
\end{sc}
\end{small}
\end{center}
\vskip 0.1in
\caption{Pricing Strategy Use of Social Game. The first two pricing strategies describe a Social Game that simulates a Demand Response competition. Flat pricing would not work for a Social Game centered around demand response instead of general energy reduction, but we put a * next to it to signify that it needs extra structure to be put into current office buildings.}
\end{table}

 \subsection{Numerical Experiments}\label{sec:methods}
 
\subsubsection{Setup}
We will now explain the pretraining procedure and how we tested it in the environment. 

To test our hypothesis that Offline-Online SAC will enable faster adaptation to unfamiliar environments like a real-world Social Game, we pretrained SAC on several simpler models of simulated person response. We then evaluated how quickly SAC, starting from the pretrained weight initialization, can learn in an OfficeLearn environment with more complex models of simulated person response. We use "Curtail and Shift" office workers in place of real office workers. We believe the transition from "Deterministic Function" workers to "Curtail and Shift" workers represents a similar step up in complexity as the transition from "Curtail and Shift" to real workers. The training environments had randomized "Deterministic Function" response types and randomized multipliers for how many "points" simulated humans received for reducing energy usage. Though the training environments used to train SAC had randomized parameters, the validation environments (with "Curtail and Shift" response types) were kept constant to ensure fairness. To ensure an accurate representation of each network's capabilities, we averaged the results from 5 different test trials and report the mean and standard error for each test. SAC is trained with an ADAM optimizer with learning rate 3e-4, 0.9 $\beta_1$, and 0.999 $\beta_2$, where $\beta_1$ refers to the first moment and $\beta_2$ refers to the second. The offline dataset used to pretrain SAC was generated with 256,000 steps from each "Deterministic Function" type environment, for a total of 768,000 state transitions, evenly distributed among the three "Deterministic Function" response models, with a variety of randomized parameters. Our intention was to provide a wide, varied, and rich dataset of simplistic responses that would allow for our Offline-Online SAC model to learn the dynamics of a Social Game without overfitting to a specific model of human response to prices. Offline-Online SAC was pretrained on this dataset for approximately 15 epochs with an ADAM optimizer with the same parameters as above.

For the planning model used in DAgger SAC, we train a 4 layer neural network, with 32 hidden units in each layer, to predict the energy usage of people throughout the day, given the prices of energy for each hour of the day. The model is trained on 1000 randomly sampled state transitions from OfficeLearn with the 'Curtail and Shift' response type. The network was trained for 10,000 epochs with the ADAM optimizer with learning rate 0.001 and L2 regularization weight 0.001. Over the 10,000 epochs, the model with the lowest loss on a holdout validation set of 256 randomly sampled transitions was used for the rest of the experiment.
 
 \begin{figure*}
\centerline{\includegraphics[width=\linewidth]{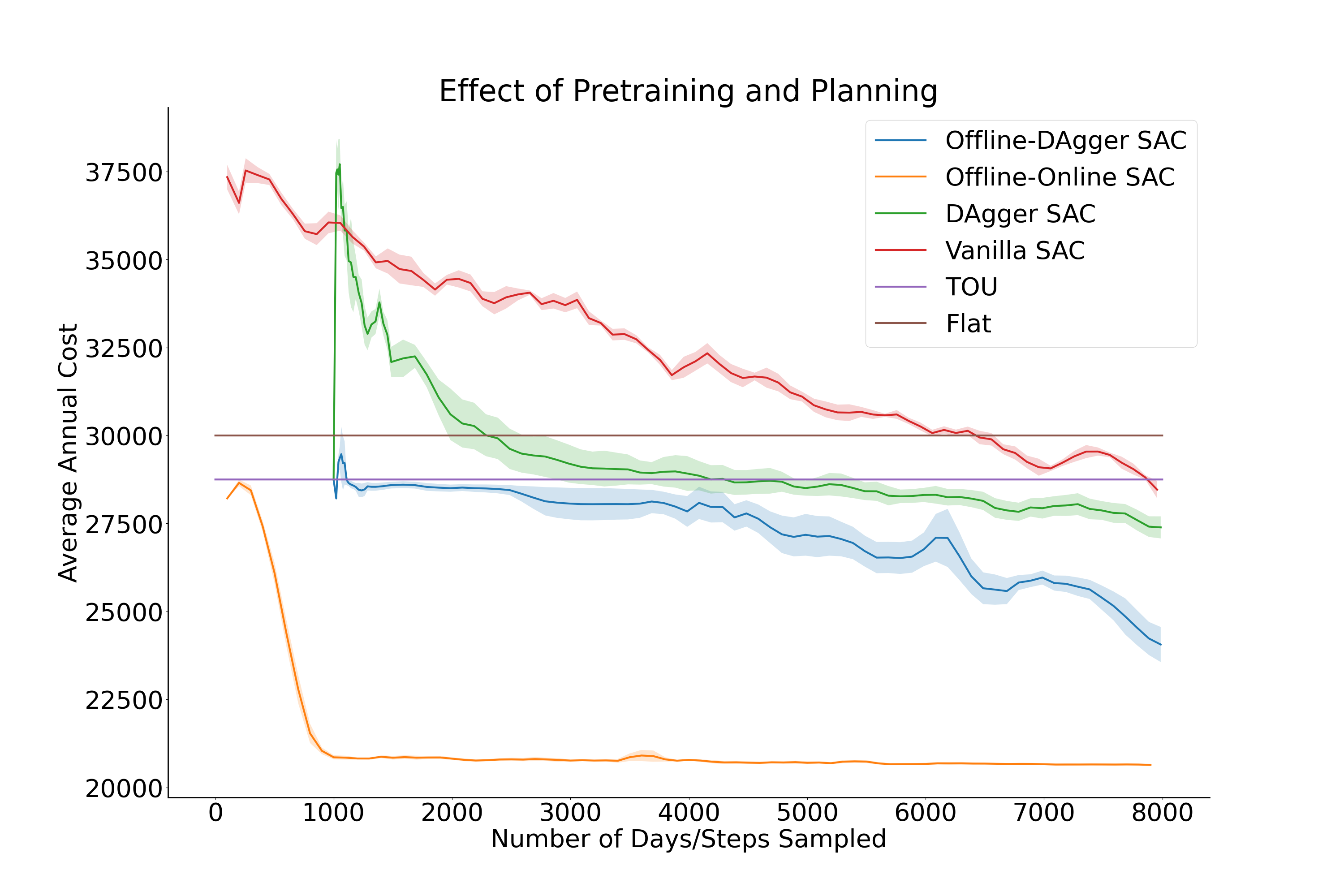}}
\caption{Offline-Online SAC and DAgger SAC Results}\label{fig:SACResults}
\begin{flushleft}
 Performance of Offline-Online SAC and DAgger SAC adapting to  "Curtail and Shift" in comparison to Online SAC. We show the mean of 5 trials, with the standard error of the mean shaded.
\end{flushleft}
\end{figure*}

 \section{Results}\label{sec:results}
 We will now describe the results obtained from our pretraining and data aggregation approaches. 
 
 \subsection{Offline-Online SAC}
 Fig. \ref{fig:SACResults} compares the performance of Offline-Online SAC, DAgger SAC, and Online SAC against our TOU and Flat Pricing baselines. In order to compare data costs, we define data cost here as the number of days' worth of data needed to train the price controller to beat the TOU baseline since it is the baseline with lowest energy cost.
 
 First, note that the costs for TOU and the RL controllers shown in the figure are inflated by \$10,000 to account for the annual cost of running the Social Game (\$400 every two weeks for a 250 day business year). If the figure shows that the RL controllers cost \$30,000 per year, \$20,000 of it is the actual cost of the energy and \$10,000 is for Social Game incentives and logistics. This inflation does not occur for the Flat Pricing baseline since it would not make sense to run the Social Game for a flat price signal. Also note that each step in the simulated Social Game represents one day.
 
 We observe that, for the first 4000 steps, Online SAC fails to beat the TOU and Flat Pricing baselines; it makes significant progress toward learning a good policy, but not enough to justify the cost of implementing it as a Social Game, even with over a simulated decade's worth of training data. Our Offline-Online SAC, however, appears to have already learned a slightly better policy than TOU during its pretraining, with an effective data cost of 0 sampled steps. In contrast, Online SAC has a data cost of 8000 days (32 years). In addition, the model converges to a price controller that appears to provide over 7000\$ in energy savings per year, with just 1000 days worth of simulated training data. The annual savings Offline-Online SAC can provide clearly justify its implementation, even with the additional cost of running the Social Game. The success of the Offline-Online SAC model also suggests that, given a dataset of simulated, more simplistic models of human behaviour, our price-setting model can learn helpful aspects of the price-setting problem that enable it to learn in a more complex environment. Our pretraining scheme appears to be robust against steps up in complexity similar to what we might encounter transitioning from the simulation to the real world.  
 
 \subsection{DAgger SAC}
 The effect of DAgger SAC is less clear cut. We plot the cost of price controllers with models trained by the planning model 1000 steps to the right, to account for the up-front data cost of 1000 steps that must be collected to train the planning model in the first place. We assume during this planning model training period TOU pricing would be used, since it is the cheapest baseline. As can be seen in Fig. \ref{fig:SACResults}, pretraining helps immensely in training the Online price controller, reducing the data cost by a factor of 2 compared to the Online controller without planning. On the other hand, the planning environment seems to slow down the training of Offline-DAgger SAC, performing only marginally better than TOU for the first 4000 days while Offline-Online SAC without the planning environment significantly diverges from TOU after just a few hundred steps.
 
\subsection{Ablation on Pretraining Steps}
Finally, we tried to observe the impact on the number of SAC pretraining steps have on our price controller's final performance in the 'Curtail and Shift' environment. We observe that even a small amount of pretraining results in a model that converges to around \$20500 in annual energy costs far quicker than Online SAC. There is some sensitivity to the number of pretraining steps on our offline dataset, appearing to cause an almost two-fold increase in required training data for full convergence in the worst case (comparing checkpoint 200 with checkpoint 400). However, even in this worst case, SAC initialized from checkpoint 200 still converges much faster than Online SAC.

\begin{figure}[h]
\centerline{\includegraphics[width=\linewidth]{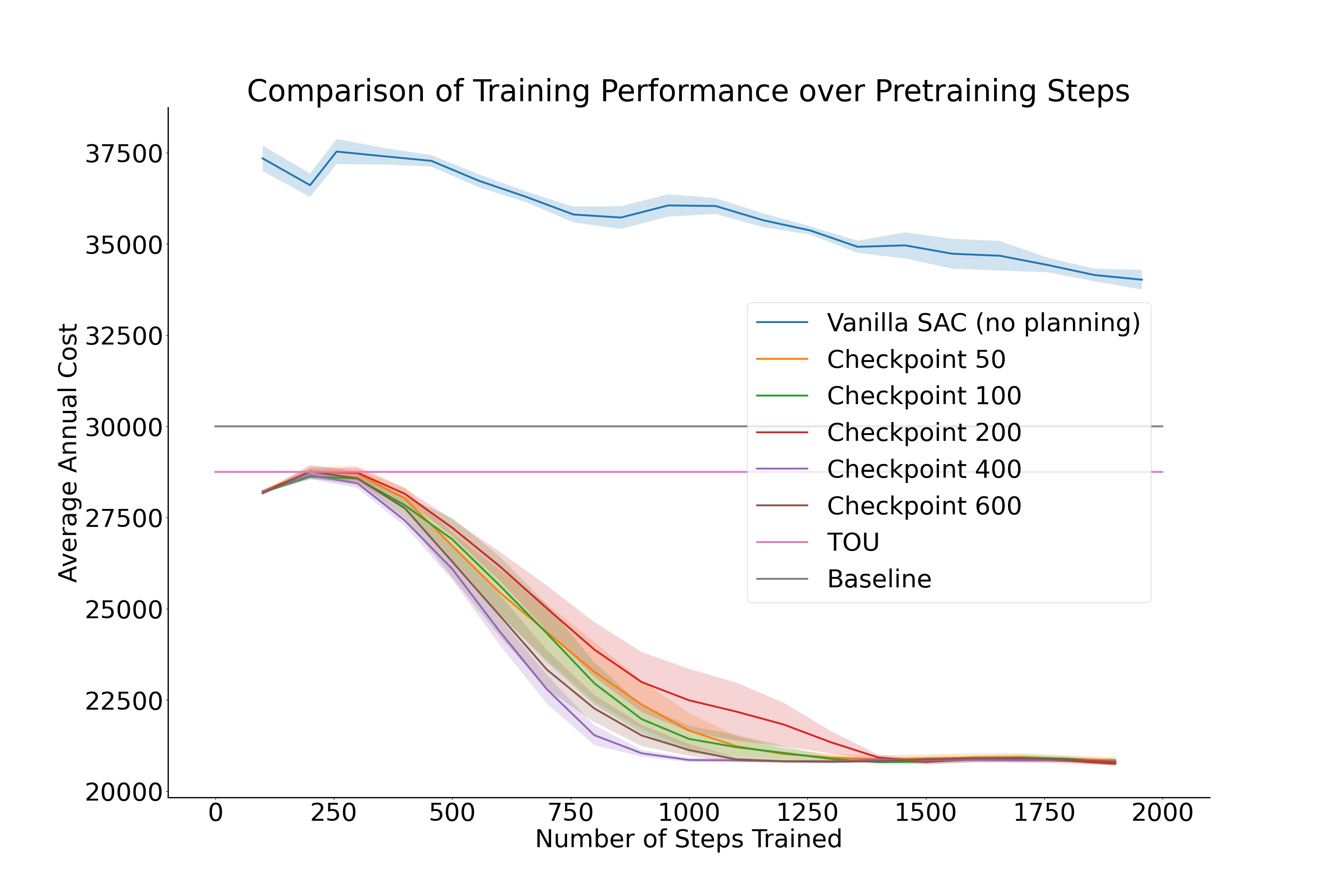}}
\caption{Pretraining Steps Ablation}\label{fig:ckptsResults}
\begin{flushleft}
 Performance of Pretrained SAC checkpoints in an environment with the  "Curtail and Shift" response type in comparison to Online SAC. We show the mean of five trials, with the standard error of the mean shaded.
\end{flushleft}
\end{figure}
 \section{Discussion}\label{sec:Discussion}
 Both our pretraining and planning approaches significantly speed up learning, but seem to be incompatible with each other. This seems to suggest that we were unable to train a planning model that performed close enough to our target environment for the pretrained controller to be aided. Curiously, the planning did not cause energy costs to increase as one might expect if training a model on a completely different task, but seemed to stabilize energy costs near that of TOU pricing while it was active. This could suggest that TOU is an appealing local minimum to the problems posed by both the target environment and the planning model.
 
 In terms of putting a price signal, our results indicate that even implementing TOU pricing produces notable cost reductions:  \$45 in savings a day for every day that the scheme is implemented. Given that our Social Game implementation requires \$400 every two weeks in incentives, implementing TOU savings would mean a savings of \$50 every two weeks. The Offline-Online SAC RL controller, meanwhile, which converged at \$78 in savings a day, can eventually reach a much larger amount of savings: roughly \$380 every two weeks. As TOU savings are relatively close to \$0, one can consider the TOU to be mostly a downpayment on investment, and the additional value from converged RL savings to be maximizing profit by many times.

A quick analysis of Californian energy's carbon intensity puts a carbon savings estimate of TOU pricing at 75\% that of the original energy consumption and 55\% of the original with a fully converged RL controller. At .52lbs CO$_2$ per kWh\cite{PGE_carbon} and ~3MWh of energy consumption every two weeks for our simulated building, this is a rough savings of 2 tons of CO$_2$ per two weeks given TOU pricing, and 3 tons of CO$_2$ per two weeks with a fully converged RL controller. 

The Online SAC controller needs over 30 years' worth of training data to converge to TOU, and thus would take many decades to recoup the initial investment in incentive for learning time. The use of the planning model in DAgger SAC speeds this up to only require 15 years to converge to TOU, but still seems to converge too slowly for timely return on investment. Since Offline-Online SAC performs better than our Flat Pricing and TOU baselines essentially immediately it seems clear that implementing this RL controller is well worth the cost of implementation.
 
 \subsection{Limitations}

It should be noted that our results perhaps under-tell the story of office demand response: we lack a structured way to measure behavior towards air conditioning, lighting, and ventilation, including a comfort model to capture the interplay between price and perception of comfort. While our simulation might ambiguously include some of these demands as generic office worker energy demands, it does not do so explicitly. Indeed, according to EIA estimates \cite{EIA_office_buildings}, lighting accounts for 17\% of energy use, ventilation 16\%, and cooling 15\%, whereas computers and office equipment, the categories that are best captured by our analysis, only account for a combined 14\%. We assume therefore that the cost estimates we provide are a lower bound on the total cost savings that a pricing scheme within an office building.

\subsection{Future Research}
Our current implementation of the planning model is not as sample-efficient as necessary for the pretrained price controller to learn from it, and even hinders its progress. Future work into creating more sample-efficient planning models may enable use of the planning model in tandem with pretraining, further accelerating training. 

Another issue with the current implementation of our planning model is that it requires collecting 1000 steps of data, during which the price controller does not learn. We did try having the price controller learn during this planning model training period, but training appeared to become very unstable once the planning model was introduced after 1000 steps (see Supplementary Material). Finding a way to ease a model trained on the real world into training with the planning model would serve to make training more efficient. 

One way this could happen is training the planning model along with the price controller and using a "reverse DAgger SAC" approach to aggregating planning model and real world data. In this case, we would distrust early planning model data, since it is unlikely to be accurate due to lack of real world data. Therefore, instead of heavily weighting the planning data at the beginning like we do in DAgger SAC, we could heavily weight the real world data in the beginning and slowly transition to more steps in the planning model as it becomes more accurate.

It would also be interesting to see if the planning model can be used for purposes other than accelerating training, e.g. increasing safety or stability. For example, providing the controller the ability to explore costly states within the planning model (like trying to incentivize usage during times of low generation instead of high) may allow the model to learn from these situations without actually having to deploy the costly changes.

\subsection{Acknowledgements}
This work is supported by the Republic of Singapore's National Research Foundation through
a grant to the Berkeley Education Alliance for Research in Singapore (BEARS) for the
Singapore–Berkeley Building Efficiency and Sustainability in the Tropics (SinBerBEST) program.


\bibliographystyle{ACM-Reference-Format}
\bibliography{sample-base}


\begin{thebibliography}{31}


\ifx \showCODEN    \undefined \def \showCODEN     #1{\unskip}     \fi
\ifx \showDOI      \undefined \def \showDOI       #1{#1}\fi
\ifx \showISBNx    \undefined \def \showISBNx     #1{\unskip}     \fi
\ifx \showISBNxiii \undefined \def \showISBNxiii  #1{\unskip}     \fi
\ifx \showISSN     \undefined \def \showISSN      #1{\unskip}     \fi
\ifx \showLCCN     \undefined \def \showLCCN      #1{\unskip}     \fi
\ifx \shownote     \undefined \def \shownote      #1{#1}          \fi
\ifx \showarticletitle \undefined \def \showarticletitle #1{#1}   \fi
\ifx \showURL      \undefined \def \showURL       {\relax}        \fi
\providecommand\bibfield[2]{#2}
\providecommand\bibinfo[2]{#2}
\providecommand\natexlab[1]{#1}
\providecommand\showeprint[2][]{arXiv:#2}

\bibitem[\protect\citeauthoryear{Agency}{Agency}{[n.d.]}]%
        {EIA_office_buildings}
\bibfield{author}{\bibinfo{person}{Energy~Information Agency}.}
  \bibinfo{year}{[n.d.]}\natexlab{}.
\newblock \showarticletitle{Energy Use Explained}.
\newblock  (\bibinfo{year}{[n.\,d.]}).
\newblock
\urldef\tempurl%
\url{https://www.eia.gov/energyexplained/use-of-energy/commercial-buildings.php}
\showURL{%
\tempurl}


\bibitem[\protect\citeauthoryear{Asadinejad, Rahimpour, Tomsovic, Qi, and
  Chen}{Asadinejad et~al\mbox{.}}{2018}]%
        {asadinejad2018evaluation}
\bibfield{author}{\bibinfo{person}{Ailin Asadinejad}, \bibinfo{person}{Alireza
  Rahimpour}, \bibinfo{person}{Kevin Tomsovic}, \bibinfo{person}{Hairong Qi},
  {and} \bibinfo{person}{Chien-fei Chen}.} \bibinfo{year}{2018}\natexlab{}.
\newblock \showarticletitle{Evaluation of residential customer elasticity for
  incentive based demand response programs}.
\newblock \bibinfo{journal}{\emph{Electric Power Systems Research}}
  \bibinfo{volume}{158} (\bibinfo{year}{2018}), \bibinfo{pages}{26--36}.
\newblock


\bibitem[\protect\citeauthoryear{Das, Konstantakopoulos, Manasawala,
  Veeravalli, Liu, and Spanos}{Das et~al\mbox{.}}{2020}]%
        {das2020occupants}
\bibfield{author}{\bibinfo{person}{Hari~Prasanna Das}, \bibinfo{person}{Ioannis
  Konstantakopoulos}, \bibinfo{person}{Aummul~Baneen Manasawala},
  \bibinfo{person}{Tanya Veeravalli}, \bibinfo{person}{Huihan Liu}, {and}
  \bibinfo{person}{Costas~J Spanos}.} \bibinfo{year}{2020}\natexlab{}.
\newblock \showarticletitle{Do Occupants in a Building exhibit patterns in
  Energy Consumption? Analyzing Clusters in Energy Social Games}.
\newblock  (\bibinfo{year}{2020}).
\newblock


\bibitem[\protect\citeauthoryear{Das, Konstantakopoulos, Manasawala,
  Veeravalli, Liu, and Spanos}{Das et~al\mbox{.}}{2019}]%
        {das2019novel}
\bibfield{author}{\bibinfo{person}{Hari~Prasanna Das},
  \bibinfo{person}{Ioannis~C Konstantakopoulos}, \bibinfo{person}{Aummul~Baneen
  Manasawala}, \bibinfo{person}{Tanya Veeravalli}, \bibinfo{person}{Huihan
  Liu}, {and} \bibinfo{person}{Costas~J Spanos}.}
  \bibinfo{year}{2019}\natexlab{}.
\newblock \showarticletitle{A novel graphical lasso based approach towards
  segmentation analysis in energy game-theoretic frameworks}. In
  \bibinfo{booktitle}{\emph{2019 18th IEEE International Conference On Machine
  Learning And Applications (ICMLA)}}. IEEE, \bibinfo{pages}{1702--1709}.
\newblock


\bibitem[\protect\citeauthoryear{Fuselli, {De Angelis}, Boaro, Squartini, Wei,
  Liu, and Piazza}{Fuselli et~al\mbox{.}}{2013}]%
        {FUSELLI2013148}
\bibfield{author}{\bibinfo{person}{Danilo Fuselli}, \bibinfo{person}{Francesco
  {De Angelis}}, \bibinfo{person}{Matteo Boaro}, \bibinfo{person}{Stefano
  Squartini}, \bibinfo{person}{Qinglai Wei}, \bibinfo{person}{Derong Liu},
  {and} \bibinfo{person}{Francesco Piazza}.} \bibinfo{year}{2013}\natexlab{}.
\newblock \showarticletitle{Action dependent heuristic dynamic programming for
  home energy resource scheduling}.
\newblock \bibinfo{journal}{\emph{International Journal of Electrical Power \&
  Energy Systems}}  \bibinfo{volume}{48} (\bibinfo{year}{2013}),
  \bibinfo{pages}{148--160}.
\newblock
\showISSN{0142-0615}
\urldef\tempurl%
\url{https://doi.org/10.1016/j.ijepes.2012.11.023}
\showDOI{\tempurl}


\bibitem[\protect\citeauthoryear{Haarnoja, Zhou, Abbeel, and Levine}{Haarnoja
  et~al\mbox{.}}{2018}]%
        {haarnoja2018soft}
\bibfield{author}{\bibinfo{person}{Tuomas Haarnoja}, \bibinfo{person}{Aurick
  Zhou}, \bibinfo{person}{Pieter Abbeel}, {and} \bibinfo{person}{Sergey
  Levine}.} \bibinfo{year}{2018}\natexlab{}.
\newblock \showarticletitle{Soft actor-critic: Off-policy maximum entropy deep
  reinforcement learning with a stochastic actor}. In
  \bibinfo{booktitle}{\emph{International Conference on Machine Learning}}.
  PMLR, \bibinfo{pages}{1861--1870}.
\newblock


\bibitem[\protect\citeauthoryear{Johnson, Starke, Abdelaziz, Jackson, and
  Tolbert}{Johnson et~al\mbox{.}}{2015}]%
        {johnson2015dynamic}
\bibfield{author}{\bibinfo{person}{Brandon~J Johnson},
  \bibinfo{person}{Michael~R Starke}, \bibinfo{person}{Omar~A Abdelaziz},
  \bibinfo{person}{Roderick~K Jackson}, {and} \bibinfo{person}{Leon~M
  Tolbert}.} \bibinfo{year}{2015}\natexlab{}.
\newblock \showarticletitle{A dynamic simulation tool for estimating demand
  response potential from residential loads}. In \bibinfo{booktitle}{\emph{2015
  IEEE Power \& Energy Society Innovative Smart Grid Technologies Conference
  (ISGT)}}. IEEE, \bibinfo{pages}{1--5}.
\newblock


\bibitem[\protect\citeauthoryear{{Kim}}{{Kim}}{2018}]%
        {8248801}
\bibfield{author}{\bibinfo{person}{Y. {Kim}}.} \bibinfo{year}{2018}\natexlab{}.
\newblock \showarticletitle{Optimal Price Based Demand Response of HVAC Systems
  in Multizone Office Buildings Considering Thermal Preferences of Individual
  Occupants Buildings}.
\newblock \bibinfo{journal}{\emph{IEEE Transactions on Industrial Informatics}}
  \bibinfo{volume}{14}, \bibinfo{number}{11} (\bibinfo{year}{2018}),
  \bibinfo{pages}{5060--5073}.
\newblock
\urldef\tempurl%
\url{https://doi.org/10.1109/TII.2018.2790429}
\showDOI{\tempurl}


\bibitem[\protect\citeauthoryear{Kofinas, Dounis, and Vouros}{Kofinas
  et~al\mbox{.}}{2018}]%
        {KOFINAS201853}
\bibfield{author}{\bibinfo{person}{P. Kofinas}, \bibinfo{person}{A.I. Dounis},
  {and} \bibinfo{person}{G.A. Vouros}.} \bibinfo{year}{2018}\natexlab{}.
\newblock \showarticletitle{Fuzzy Q-Learning for multi-agent decentralized
  energy management in microgrids}.
\newblock \bibinfo{journal}{\emph{Applied Energy}}  \bibinfo{volume}{219}
  (\bibinfo{year}{2018}), \bibinfo{pages}{53--67}.
\newblock
\showISSN{0306-2619}
\urldef\tempurl%
\url{https://doi.org/10.1016/j.apenergy.2018.03.017}
\showDOI{\tempurl}


\bibitem[\protect\citeauthoryear{Konstantakopoulos, Barkan, He, Veeravalli,
  Liu, and Spanos}{Konstantakopoulos et~al\mbox{.}}{2019a}]%
        {konstantakopoulos2019deep}
\bibfield{author}{\bibinfo{person}{Ioannis~C Konstantakopoulos},
  \bibinfo{person}{Andrew~R Barkan}, \bibinfo{person}{Shiying He},
  \bibinfo{person}{Tanya Veeravalli}, \bibinfo{person}{Huihan Liu}, {and}
  \bibinfo{person}{Costas Spanos}.} \bibinfo{year}{2019}\natexlab{a}.
\newblock \showarticletitle{A deep learning and gamification approach to
  improving human-building interaction and energy efficiency in smart
  infrastructure}.
\newblock \bibinfo{journal}{\emph{Applied energy}}  \bibinfo{volume}{237}
  (\bibinfo{year}{2019}), \bibinfo{pages}{810--821}.
\newblock


\bibitem[\protect\citeauthoryear{Konstantakopoulos, Das, Barkan, He,
  Veeravalli, Liu, Manasawala, Lin, and Spanos}{Konstantakopoulos
  et~al\mbox{.}}{2019b}]%
        {konstantakopoulos2019design}
\bibfield{author}{\bibinfo{person}{Ioannis~C Konstantakopoulos},
  \bibinfo{person}{Hari~Prasanna Das}, \bibinfo{person}{Andrew~R Barkan},
  \bibinfo{person}{Shiying He}, \bibinfo{person}{Tanya Veeravalli},
  \bibinfo{person}{Huihan Liu}, \bibinfo{person}{Aummul~Baneen Manasawala},
  \bibinfo{person}{Yu-Wen Lin}, {and} \bibinfo{person}{Costas~J Spanos}.}
  \bibinfo{year}{2019}\natexlab{b}.
\newblock \showarticletitle{Design, benchmarking and explainability analysis of
  a game-theoretic framework towards energy efficiency in smart
  infrastructure}.
\newblock \bibinfo{journal}{\emph{arXiv preprint arXiv:1910.07899}}
  (\bibinfo{year}{2019}).
\newblock


\bibitem[\protect\citeauthoryear{Konstantakopoulos, Ratliff, Jin, Sastry, and
  Spanos}{Konstantakopoulos et~al\mbox{.}}{2017}]%
        {konstantakopoulos2017robust}
\bibfield{author}{\bibinfo{person}{Ioannis~C Konstantakopoulos},
  \bibinfo{person}{Lillian~J Ratliff}, \bibinfo{person}{Ming Jin},
  \bibinfo{person}{S~Shankar Sastry}, {and} \bibinfo{person}{Costas~J Spanos}.}
  \bibinfo{year}{2017}\natexlab{}.
\newblock \showarticletitle{A robust utility learning framework via inverse
  optimization}.
\newblock \bibinfo{journal}{\emph{IEEE Transactions on Control Systems
  Technology}} \bibinfo{volume}{26}, \bibinfo{number}{3}
  (\bibinfo{year}{2017}), \bibinfo{pages}{954--970}.
\newblock


\bibitem[\protect\citeauthoryear{Li, Liu, Yu, Deng, Huang, and Liu}{Li
  et~al\mbox{.}}{2018}]%
        {li2018integrating}
\bibfield{author}{\bibinfo{person}{Chaojie Li}, \bibinfo{person}{Chen Liu},
  \bibinfo{person}{Xinghuo Yu}, \bibinfo{person}{Ke Deng},
  \bibinfo{person}{Tingwen Huang}, {and} \bibinfo{person}{Liangchen Liu}.}
  \bibinfo{year}{2018}\natexlab{}.
\newblock \showarticletitle{Integrating Demand Response and Renewable Energy In
  Wholesale Market.}. In \bibinfo{booktitle}{\emph{IJCAI}}.
  \bibinfo{pages}{382--388}.
\newblock


\bibitem[\protect\citeauthoryear{{Li} and {Jayaweera}}{{Li} and
  {Jayaweera}}{2014}]%
        {7018632}
\bibfield{author}{\bibinfo{person}{D. {Li}} {and} \bibinfo{person}{S.~K.
  {Jayaweera}}.} \bibinfo{year}{2014}\natexlab{}.
\newblock \showarticletitle{Reinforcement learning aided smart-home
  decision-making in an interactive smart grid}. In
  \bibinfo{booktitle}{\emph{2014 IEEE Green Energy and Systems Conference
  (IGESC)}}. \bibinfo{pages}{1--6}.
\newblock
\urldef\tempurl%
\url{https://doi.org/10.1109/IGESC.2014.7018632}
\showDOI{\tempurl}


\bibitem[\protect\citeauthoryear{Liang, Liaw, Nishihara, Moritz, Fox, Goldberg,
  Gonzalez, Jordan, and Stoica}{Liang et~al\mbox{.}}{2018}]%
        {liang2018rllib}
\bibfield{author}{\bibinfo{person}{Eric Liang}, \bibinfo{person}{Richard Liaw},
  \bibinfo{person}{Robert Nishihara}, \bibinfo{person}{Philipp Moritz},
  \bibinfo{person}{Roy Fox}, \bibinfo{person}{Ken Goldberg},
  \bibinfo{person}{Joseph Gonzalez}, \bibinfo{person}{Michael Jordan}, {and}
  \bibinfo{person}{Ion Stoica}.} \bibinfo{year}{2018}\natexlab{}.
\newblock \showarticletitle{RLlib: Abstractions for distributed reinforcement
  learning}. In \bibinfo{booktitle}{\emph{International Conference on Machine
  Learning}}. PMLR, \bibinfo{pages}{3053--3062}.
\newblock


\bibitem[\protect\citeauthoryear{{Liu}, {Yuen}, {Ul Hassan}, {Huang}, {Yu}, and
  {Xie}}{{Liu} et~al\mbox{.}}{2015}]%
        {6963416}
\bibfield{author}{\bibinfo{person}{Y. {Liu}}, \bibinfo{person}{C. {Yuen}},
  \bibinfo{person}{N. {Ul Hassan}}, \bibinfo{person}{S. {Huang}},
  \bibinfo{person}{R. {Yu}}, {and} \bibinfo{person}{S. {Xie}}.}
  \bibinfo{year}{2015}\natexlab{}.
\newblock \showarticletitle{Electricity Cost Minimization for a Microgrid With
  Distributed Energy Resource Under Different Information Availability}.
\newblock \bibinfo{journal}{\emph{IEEE Transactions on Industrial Electronics}}
  \bibinfo{volume}{62}, \bibinfo{number}{4} (\bibinfo{year}{2015}),
  \bibinfo{pages}{2571--2583}.
\newblock
\urldef\tempurl%
\url{https://doi.org/10.1109/TIE.2014.2371780}
\showDOI{\tempurl}


\bibitem[\protect\citeauthoryear{Ma, Hu, and Spanos}{Ma et~al\mbox{.}}{2015}]%
        {ma2015cooperative}
\bibfield{author}{\bibinfo{person}{Kai Ma}, \bibinfo{person}{Guoqiang Hu},
  {and} \bibinfo{person}{Costas~J Spanos}.} \bibinfo{year}{2015}\natexlab{}.
\newblock \showarticletitle{A cooperative demand response scheme using
  punishment mechanism and application to industrial refrigerated warehouses}.
\newblock \bibinfo{journal}{\emph{IEEE Transactions on Industrial Informatics}}
  \bibinfo{volume}{11}, \bibinfo{number}{6} (\bibinfo{year}{2015}),
  \bibinfo{pages}{1520--1531}.
\newblock


\bibitem[\protect\citeauthoryear{Mbuwir, Ruelens, Spiessens, and
  Deconinck}{Mbuwir et~al\mbox{.}}{2017}]%
        {en10111846}
\bibfield{author}{\bibinfo{person}{Brida~V. Mbuwir}, \bibinfo{person}{Frederik
  Ruelens}, \bibinfo{person}{Fred Spiessens}, {and} \bibinfo{person}{Geert
  Deconinck}.} \bibinfo{year}{2017}\natexlab{}.
\newblock \showarticletitle{Battery Energy Management in a Microgrid Using
  Batch Reinforcement Learning}.
\newblock \bibinfo{journal}{\emph{Energies}} \bibinfo{volume}{10},
  \bibinfo{number}{11} (\bibinfo{year}{2017}).
\newblock
\showISSN{1996-1073}
\urldef\tempurl%
\url{https://www.mdpi.com/1996-1073/10/11/1846}
\showURL{%
\tempurl}


\bibitem[\protect\citeauthoryear{Mnih, Kavukcuoglu, Silver, Graves, Antonoglou,
  Wierstra, and Riedmiller}{Mnih et~al\mbox{.}}{2013}]%
        {mnih2013playing}
\bibfield{author}{\bibinfo{person}{Volodymyr Mnih}, \bibinfo{person}{Koray
  Kavukcuoglu}, \bibinfo{person}{David Silver}, \bibinfo{person}{Alex Graves},
  \bibinfo{person}{Ioannis Antonoglou}, \bibinfo{person}{Daan Wierstra}, {and}
  \bibinfo{person}{Martin Riedmiller}.} \bibinfo{year}{2013}\natexlab{}.
\newblock \showarticletitle{Playing atari with deep reinforcement learning}.
\newblock \bibinfo{journal}{\emph{arXiv preprint arXiv:1312.5602}}
  (\bibinfo{year}{2013}).
\newblock


\bibitem[\protect\citeauthoryear{PGE}{PGE}{[n.d.]}]%
        {PGE_carbon}
\bibfield{author}{\bibinfo{person}{PGE}.} \bibinfo{year}{[n.d.]}\natexlab{}.
\newblock \showarticletitle{Pacific Gas and Electric Company Carbon Footprint
  Calculator Assumptions}.
\newblock  (\bibinfo{year}{[n.\,d.]}).
\newblock
\urldef\tempurl%
\url{https://www.pge.com/includes/docs/pdfs/about/environment/calculator/assumptions.pdf}
\showURL{%
\tempurl}


\bibitem[\protect\citeauthoryear{Raju, Sankar, and Milton}{Raju
  et~al\mbox{.}}{2015}]%
        {RAJU2015231}
\bibfield{author}{\bibinfo{person}{Leo Raju}, \bibinfo{person}{Sibi Sankar},
  {and} \bibinfo{person}{R.S. Milton}.} \bibinfo{year}{2015}\natexlab{}.
\newblock \showarticletitle{Distributed Optimization of Solar Micro-grid Using
  Multi Agent Reinforcement Learning}.
\newblock \bibinfo{journal}{\emph{Procedia Computer Science}}
  \bibinfo{volume}{46} (\bibinfo{year}{2015}), \bibinfo{pages}{231--239}.
\newblock
\showISSN{1877-0509}
\urldef\tempurl%
\url{https://doi.org/10.1016/j.procs.2015.02.016}
\showDOI{\tempurl}
\newblock
\shownote{Proceedings of the International Conference on Information and
  Communication Technologies, ICICT 2014, 3-5 December 2014 at Bolgatty Palace
  \& Island Resort, Kochi, India.}


\bibitem[\protect\citeauthoryear{Ross, Gordon, and Bagnell}{Ross
  et~al\mbox{.}}{2011}]%
        {ross2011reduction}
\bibfield{author}{\bibinfo{person}{St{\'e}phane Ross},
  \bibinfo{person}{Geoffrey Gordon}, {and} \bibinfo{person}{Drew Bagnell}.}
  \bibinfo{year}{2011}\natexlab{}.
\newblock \showarticletitle{A reduction of imitation learning and structured
  prediction to no-regret online learning}. In
  \bibinfo{booktitle}{\emph{Proceedings of the fourteenth international
  conference on artificial intelligence and statistics}}. JMLR Workshop and
  Conference Proceedings, \bibinfo{pages}{627--635}.
\newblock


\bibitem[\protect\citeauthoryear{Spangher, Gokul, Khattar, Palakapilly, Agwan,
  Tawade, and Spanos}{Spangher et~al\mbox{.}}{2020a}]%
        {spangher2020augmenting}
\bibfield{author}{\bibinfo{person}{Lucas Spangher}, \bibinfo{person}{Akash
  Gokul}, \bibinfo{person}{Manan Khattar}, \bibinfo{person}{Joseph
  Palakapilly}, \bibinfo{person}{Utkarsha Agwan}, \bibinfo{person}{Akaash
  Tawade}, {and} \bibinfo{person}{Costas Spanos}.}
  \bibinfo{year}{2020}\natexlab{a}.
\newblock \showarticletitle{Augmenting Reinforcement Learning with a Planning
  Model for Optimizing Energy Demand Response}. In
  \bibinfo{booktitle}{\emph{Proceedings of the 1st International Workshop on
  Reinforcement Learning for Energy Management in Buildings \& Cities}}.
  \bibinfo{pages}{39--42}.
\newblock


\bibitem[\protect\citeauthoryear{Spangher, Gokul, Khattar, Palakapilly, Tawade,
  Bouyamourn, Devonport, and Spanos}{Spangher et~al\mbox{.}}{2020b}]%
        {spangher2020prospective}
\bibfield{author}{\bibinfo{person}{Lucas Spangher}, \bibinfo{person}{Akash
  Gokul}, \bibinfo{person}{Manan Khattar}, \bibinfo{person}{Joseph
  Palakapilly}, \bibinfo{person}{Akaash Tawade}, \bibinfo{person}{Adam
  Bouyamourn}, \bibinfo{person}{Alex Devonport}, {and} \bibinfo{person}{Costas
  Spanos}.} \bibinfo{year}{2020}\natexlab{b}.
\newblock \showarticletitle{Prospective experiment for reinforcement learning
  on demand response in a social game framework}. In
  \bibinfo{booktitle}{\emph{Proceedings of the Eleventh ACM International
  Conference on Future Energy Systems}}. \bibinfo{pages}{438--444}.
\newblock


\bibitem[\protect\citeauthoryear{Spangher, Gokul, Palakapilly, Agwan, Khattar,
  Ma, and Spanos}{Spangher et~al\mbox{.}}{[n.d.]}]%
        {spangherofficelearn}
\bibfield{author}{\bibinfo{person}{Lucas Spangher}, \bibinfo{person}{Akash
  Gokul}, \bibinfo{person}{Joseph Palakapilly}, \bibinfo{person}{Utkarsha
  Agwan}, \bibinfo{person}{Manan Khattar}, \bibinfo{person}{Wann-Jiun Ma},
  {and} \bibinfo{person}{Costas Spanos}.} \bibinfo{year}{[n.d.]}\natexlab{}.
\newblock \showarticletitle{OfficeLearn: An OpenAI Gym Environment for
  Reinforcement Learning on Occupant-Level Building’s Energy Demand
  Response}. In \bibinfo{booktitle}{\emph{Tackling Climate Change with
  Artificial Intelligence Workshop at NeurIPS, 2020}}.
\newblock


\bibitem[\protect\citeauthoryear{Spangher, Tawade, Devonport, and
  Spanos}{Spangher et~al\mbox{.}}{2019}]%
        {spangher2019visualization}
\bibfield{author}{\bibinfo{person}{Lucas Spangher}, \bibinfo{person}{Akaash
  Tawade}, \bibinfo{person}{Alex Devonport}, {and} \bibinfo{person}{Costas
  Spanos}.} \bibinfo{year}{2019}\natexlab{}.
\newblock \showarticletitle{Engineering vs. Ambient Type Visualizations:
  Quantifying Effects of Different Data Visualizations on Energy Consumption}.
  In \bibinfo{booktitle}{\emph{Proceedings of the 1st ACM International
  Workshop on Urban Building Energy Sensing, Controls, Big Data Analysis, and
  Visualization}} (New York, NY, USA) \emph{(\bibinfo{series}{UrbSys'19})}.
  \bibinfo{publisher}{Association for Computing Machinery},
  \bibinfo{address}{New York, NY, USA}, \bibinfo{pages}{14–22}.
\newblock
\showISBNx{9781450370141}
\urldef\tempurl%
\url{https://doi.org/10.1145/3363459.3363527}
\showDOI{\tempurl}


\bibitem[\protect\citeauthoryear{Sutton and Barto}{Sutton and Barto}{2018}]%
        {sutton2018reinforcement}
\bibfield{author}{\bibinfo{person}{Richard~S Sutton} {and}
  \bibinfo{person}{Andrew~G Barto}.} \bibinfo{year}{2018}\natexlab{}.
\newblock \bibinfo{booktitle}{\emph{Reinforcement learning: An introduction}}.
\newblock \bibinfo{publisher}{MIT press}.
\newblock


\bibitem[\protect\citeauthoryear{Tesauro}{Tesauro}{1994}]%
        {tesauro1994td}
\bibfield{author}{\bibinfo{person}{Gerald Tesauro}.}
  \bibinfo{year}{1994}\natexlab{}.
\newblock \showarticletitle{TD-Gammon, a self-teaching backgammon program,
  achieves master-level play}.
\newblock \bibinfo{journal}{\emph{Neural computation}} \bibinfo{volume}{6},
  \bibinfo{number}{2} (\bibinfo{year}{1994}), \bibinfo{pages}{215--219}.
\newblock


\bibitem[\protect\citeauthoryear{{Wei}, {Liu}, and {Shi}}{{Wei}
  et~al\mbox{.}}{2015}]%
        {6915886}
\bibfield{author}{\bibinfo{person}{Q. {Wei}}, \bibinfo{person}{D. {Liu}}, {and}
  \bibinfo{person}{G. {Shi}}.} \bibinfo{year}{2015}\natexlab{}.
\newblock \showarticletitle{A novel dual iterative Q-learning method for
  optimal battery management in smart residential environments}.
\newblock \bibinfo{journal}{\emph{IEEE Transactions on Industrial Electronics}}
  \bibinfo{volume}{62}, \bibinfo{number}{4} (\bibinfo{year}{2015}),
  \bibinfo{pages}{2509--2518}.
\newblock
\urldef\tempurl%
\url{https://doi.org/10.1109/TIE.2014.2361485}
\showDOI{\tempurl}


\bibitem[\protect\citeauthoryear{Yoon, Baldick, and Novoselac}{Yoon
  et~al\mbox{.}}{2014}]%
        {yoon2014dynamic}
\bibfield{author}{\bibinfo{person}{Ji~Hoon Yoon}, \bibinfo{person}{Ross
  Baldick}, {and} \bibinfo{person}{Atila Novoselac}.}
  \bibinfo{year}{2014}\natexlab{}.
\newblock \showarticletitle{Dynamic demand response controller based on
  real-time retail price for residential buildings}.
\newblock \bibinfo{journal}{\emph{IEEE Transactions on Smart Grid}}
  \bibinfo{volume}{5}, \bibinfo{number}{1} (\bibinfo{year}{2014}),
  \bibinfo{pages}{121--129}.
\newblock


\bibitem[\protect\citeauthoryear{{Zhang} and {van der Schaar}}{{Zhang} and {van
  der Schaar}}{2014}]%
        {6848212}
\bibfield{author}{\bibinfo{person}{Y. {Zhang}} {and} \bibinfo{person}{M. {van
  der Schaar}}.} \bibinfo{year}{2014}\natexlab{}.
\newblock \showarticletitle{Structure-aware stochastic load management in smart
  grids}. In \bibinfo{booktitle}{\emph{IEEE INFOCOM 2014 - IEEE Conference on
  Computer Communications}}. \bibinfo{pages}{2643--2651}.
\newblock
\urldef\tempurl%
\url{https://doi.org/10.1109/INFOCOM.2014.6848212}
\showDOI{\tempurl}


\end{thebibliography}

\section{Supplementary Material}
\subsection{2-Stage DAgger}
We also tested a variant of our planning model integration approach where, instead of starting the price controller's training after 1000 steps, we start training immediately (at 0 steps). We then transition to our previously described planning model/target environment alternation after the planning model has been trained after 1000 steps. In doing so, we sought to take advantage of those first 1000 steps for training the price controller, to increase overall sample efficiency. We named this variant "2 Stage DAgger SAC", where the first stage would be the controller training as normal for 1000 steps, and the second would be training with our planning scheme starting from 1000 steps.

\begin{figure}[t]
\centerline{\includegraphics[width=\linewidth]{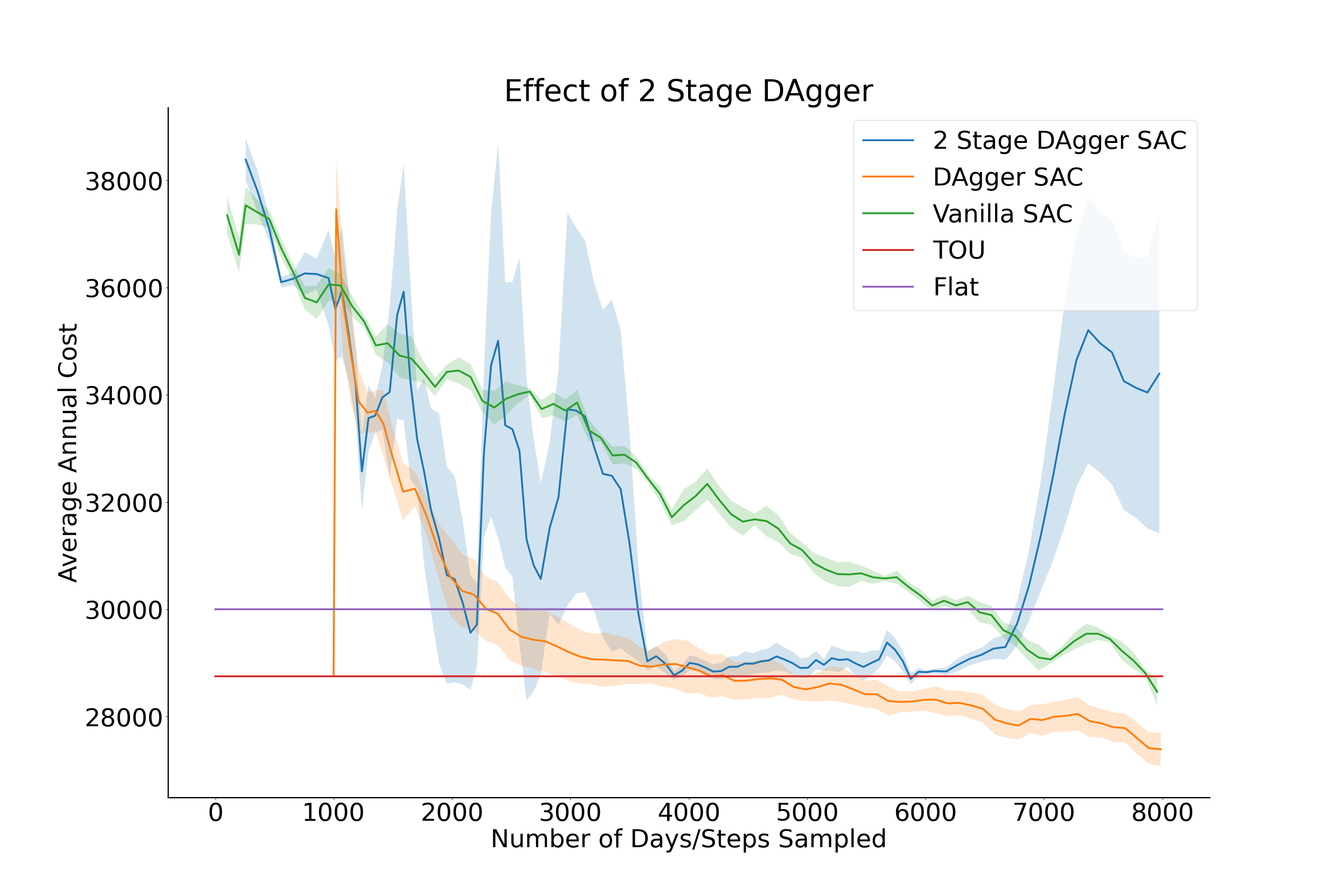}}
\caption{2 Stage DAgger Results}\label{fig:2StageResults}
\begin{flushleft}
 Performance of 2 Stage DAgger training in an environment with the  "Curtail and Shift" response type in comparison to Online SAC and our previously described planning model. We show the mean of five trials, with the standard error of the mean shaded.
\end{flushleft}
\end{figure}

As can be seen in Fig. \ref{fig:2StageResults}, however, this does not appear to have had a positive effect on training, serving only to increase instability in training. Although it does converge to a cost similar to TOU faster than Online SAC, the 2 Stage approach's energy costs began to spike instead of decrease, at around 7000 steps. It is likely that the sudden transition from $100\%$ target environment steps in SAC's replay buffer to $\sim 91\%$ planning model steps at step 1000 confused the price controller model, resulting in the observed instability.

\end{document}